\documentclass{article}

\usepackage[preprint]{neurips_2026}
\usepackage[utf8]{inputenc}
\usepackage[T1]{fontenc}
\usepackage{graphicx}
\usepackage{amsmath}
\usepackage{amssymb}
\usepackage{amsfonts}
\usepackage{booktabs}
\usepackage{multirow}
\usepackage{array}
\usepackage{enumitem}
\usepackage{microtype}
\usepackage{xcolor}
\usepackage{url}
\usepackage{xspace}
\usepackage{fancyhdr}
\usepackage{algorithm}
\usepackage{algpseudocode}
\usepackage{hyperref}

\hypersetup{
    colorlinks=true,
    citecolor=blue!60!black,
    urlcolor=blue!60!black,
    linkcolor=blue!60!black
}

\newcommand{\githubicon}{\raisebox{-1.5pt}{\includegraphics[height=1.05em]{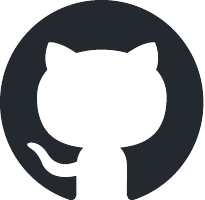}}}
\newcommand{\hficon}{\raisebox{-1.5pt}{\includegraphics[height=1.05em]{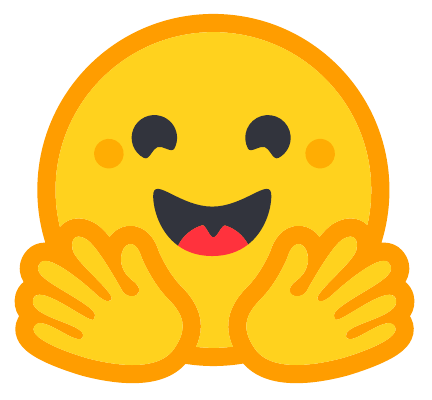}}}
\newcommand{\policy}{\pi_{\theta}}
\newcommand{\harness}{H}
\newcommand{\adapter}{A}
\newcommand{\dom}{\mathrm{dom}}
\newcommand{\old}{\mathrm{old}}
\newcommand{\new}{\mathrm{new}}

\title{\textbf{Dual-Loop Agent Specialization: Co-Adapting Policy Weights and Harness Workspace}}

\author{%
BoHao Chen\\
Independent Researcher\\
\texttt{chenbh.cs@gmail.com}\\[0.6em]
\githubicon\;\href{https://github.com/bhaochen/harvo}{\texttt{github.com/bhaochen/harvo}}\\
\hficon\;\href{https://huggingface.co/bhaochen/harvo}{\texttt{huggingface.co/bhaochen/harvo}}\\
}

\fancypagestyle{firstpage}{%
  \fancyhf{}%
  \fancyhead[R]{}%
}

\begin{document}
\maketitle
\thispagestyle{firstpage}

\begin{abstract}
Domain specialization of LLM agents is often framed as a choice between updating model weights and editing the surrounding agent harness. We instead study them as complementary online loops over a shared task-level rollout pattern. The first loop is \textbf{SAO RL}: we engineer a real-time single-rollout, full-parameter training path for long-horizon agent traces, with turn-level SAO credit, action-token-only gradients, exact streaming over 200k contexts, fused critic/actor objectives, prompt-prefix reuse, and restart-based quantized vLLM serving. The second loop evolves the \textbf{harness workspace}---the system prompt, tool registry, skills, exemplars, and manifest---with Evo-Harness, which turns one-shot execution contexts into auditable workspace edits through grounded reflection. Weight updates change the model's conditional distribution; harness evolution changes the evidence, tools, and procedures supplied to that model. Neither loop is treated as a substitute for the other. In a six-instance SWE-bench Verified training study with Codev, a Claude Code-style coding agent, the SAO loop resolves all six rollouts while the harness loop resolves four, with one verifier failure and one 3{,}000-second timeout. Their mean process rewards round to the same value (0.414) but have different temporal profiles. The final Evo artifact is one system-prompt edit produced after three reflection attempts on the same six rollouts. We audit solve rates, process signals, token accounting, and streamed reasoning lengths from these completed runs, and find that reasoning lengths are not uniformly compressed. These results motivate vertical specialization as joint adaptation of policy weights and harness workspace rather than a competition between them; cross-domain, combined-configuration, and retention-gate evaluation remain future work.
\end{abstract}

\section{Introduction}

\textbf{Agents are harness plus policy.} A modern LLM agent is not just a model. It is a \emph{policy} $\policy$---the language model with weights $\theta$---embedded in a \emph{harness} $\harness$ that owns the transcript, exposes tools, injects skills, manages context, and decides when to plan or reflect \citep{yang2024sweagent,yao2023react}. The same policy can behave very differently under two harnesses: a terminal harness with a single bash tool encourages terse, executable actions, while an office harness with document APIs encourages multi-step UI automation. Most specialization pipelines optimize only $\theta$ or only $\harness$ and leave the other object hand-designed; we study how both can be adapted from the same agent-environment interaction.

\begin{figure}[t]
    \centering
    \includegraphics[width=0.92\linewidth]{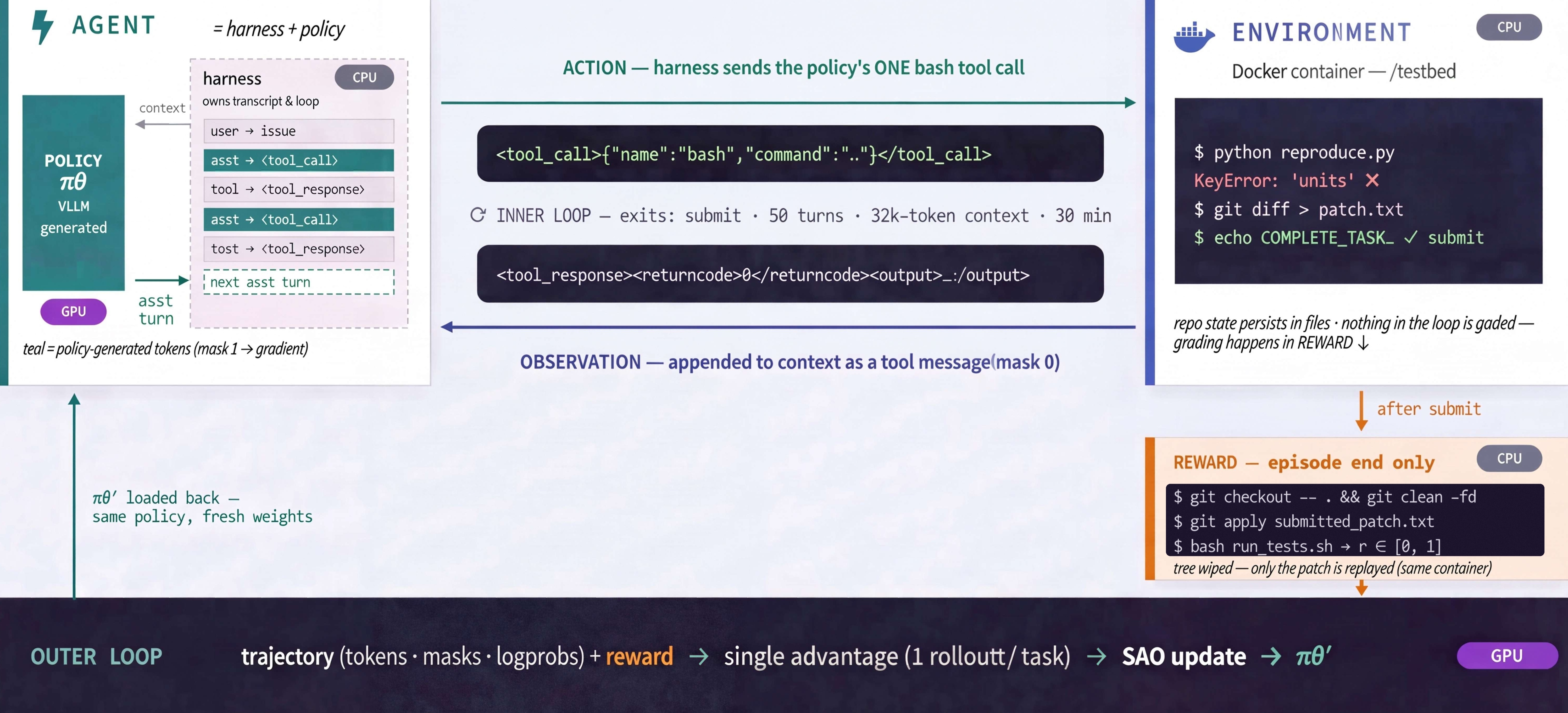}
    \caption{Weight-loop specialization. The harness sends the policy's tool call to a Docker environment; the environment returns an observation and, at episode end, a scalar reward. The outer loop updates the full policy weights with SAO.}
    \label{fig:rl-loop}
\end{figure}

\begin{figure}[t]
    \centering
    \includegraphics[width=0.92\linewidth]{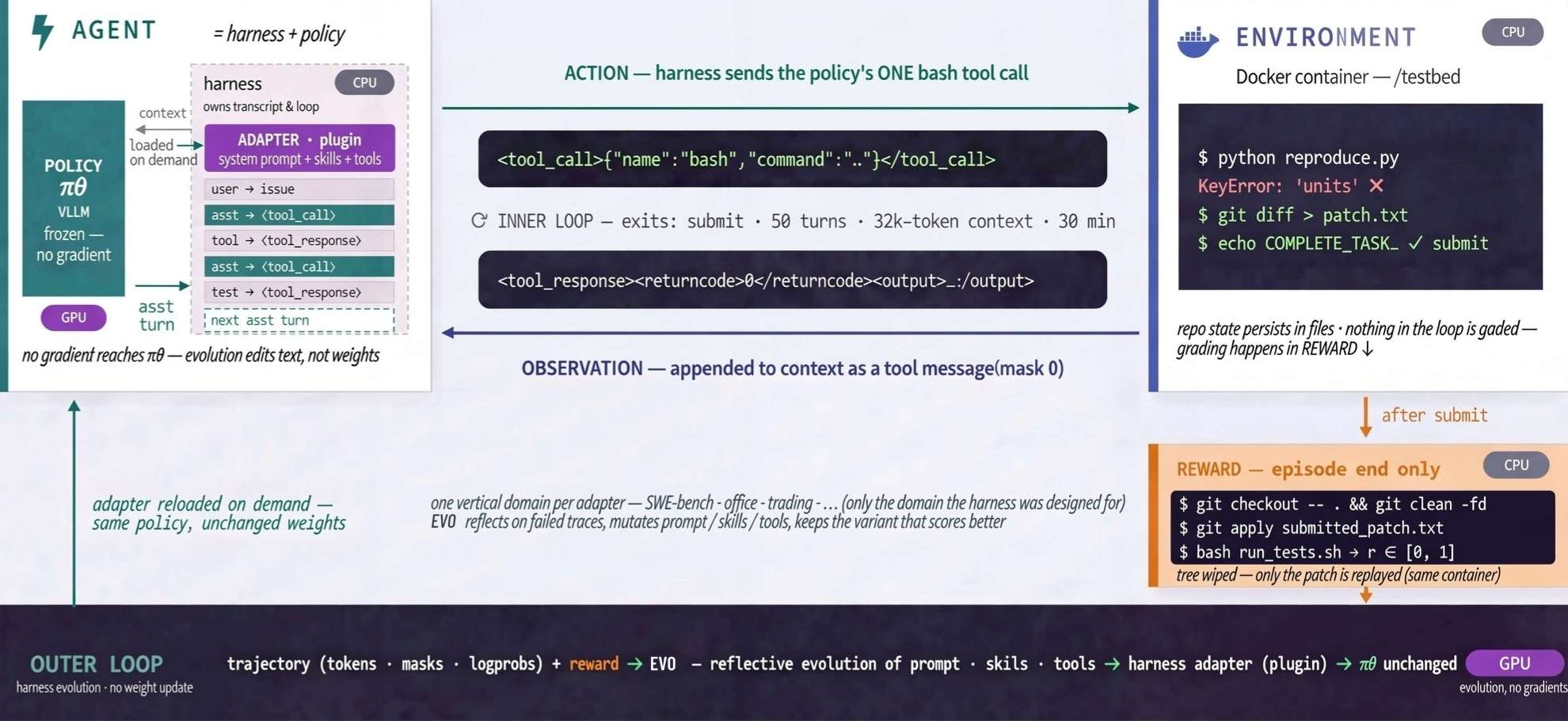}
    \caption{Harness-loop specialization. The inner loop follows the same terminal-agent interaction pattern as weight-loop specialization, but the outer loop turns execution contexts into workspace-file edits with Evo-Harness. The policy is frozen when this loop runs alone; the resulting workspace could be reloaded on demand per vertical domain and combined with a trained policy.}
    \label{fig:harness-evolution}
\end{figure}

\textbf{Two loops over one rollout stream.} Weight-based specialization (Figure~\ref{fig:rl-loop}) is powerful but demands optimizer state, GPU time, and infrastructure for long-horizon trajectories. Harness evolution (Figure~\ref{fig:harness-evolution}) demands no weight-training step and instead edits the reusable scaffolding around a policy. These are not substitutes: one changes the conditional distribution inside the model, while the other changes the information and procedures supplied to the model. Our completed SAO run resolves all six assigned coding tasks, while the frozen-policy harness run resolves four of six, with one verifier failure and one timeout. The two mean process rewards round to 0.414 on both sides, but their per-update and per-task profiles differ; streamed reasoning-length medians are 342 and 506 characters (Figures~\ref{fig:loop-profiles} and \ref{fig:length}). We do not instrument truncation or format rescue in these runs, so we make no claim about an SFT-specific robustness effect. Instead, we treat both paths as first-class specialization mechanisms and audit what each one changes: the serving, capture, and streaming machinery required for real-time SAO on long traces, and the evidence-to-workspace compilation path required for one-shot harness evolution.

\textbf{Harness evolution.} We also treat the harness workspace as an object of optimization. An outer loop observes one-shot execution contexts, reflects on success and failure evidence, and applies auditable edits to workspace files such as prompts, tool descriptions, skills, exemplars, and planning templates. The inner loop is unchanged (Figure~\ref{fig:harness-evolution}); when this loop runs alone, $\theta$ remains frozen. The output is a specialized workspace that can be materialized as a \emph{harness adapter} for deployment and loaded on demand for a vertical domain. This is a deployment-level analogy to LoRA \citep{hu2021lora} and AdapterFusion \citep{pfeiffer2021adapterfusion}: small, composable, reversible deltas rather than a new monolithic checkpoint. The analogy is not mathematical equivalence; a harness adapter is a file-level delta, not a parameter-space update or a learned fusion module.

\textbf{Why this can work.} Weight updates and harness edits attack different bottlenecks. RL solves ``the policy can do this but does it poorly'': it sharpens an existing mode through on-policy reward. Harness evolution solves ``the policy has the capability but the scaffolding elicits the wrong behavior'': it changes the conditional distribution of actions without changing the policy's support. The search space is also different. A harness has discrete, interpretable components with strong structural priors; a grounded reflection can propose a reusable procedure such as ``always run the failing test before editing'' or ``use \texttt{grep -R} before opening files.'' Evo-Harness exploits this property by compiling execution evidence into workspace-file edits, creating a new skill only when no existing file can carry the lesson \citep{wei2026evoharnesscontexttoharnessskillcompilation}.

\begin{figure}[t]
    \centering
    \includegraphics[width=0.88\linewidth]{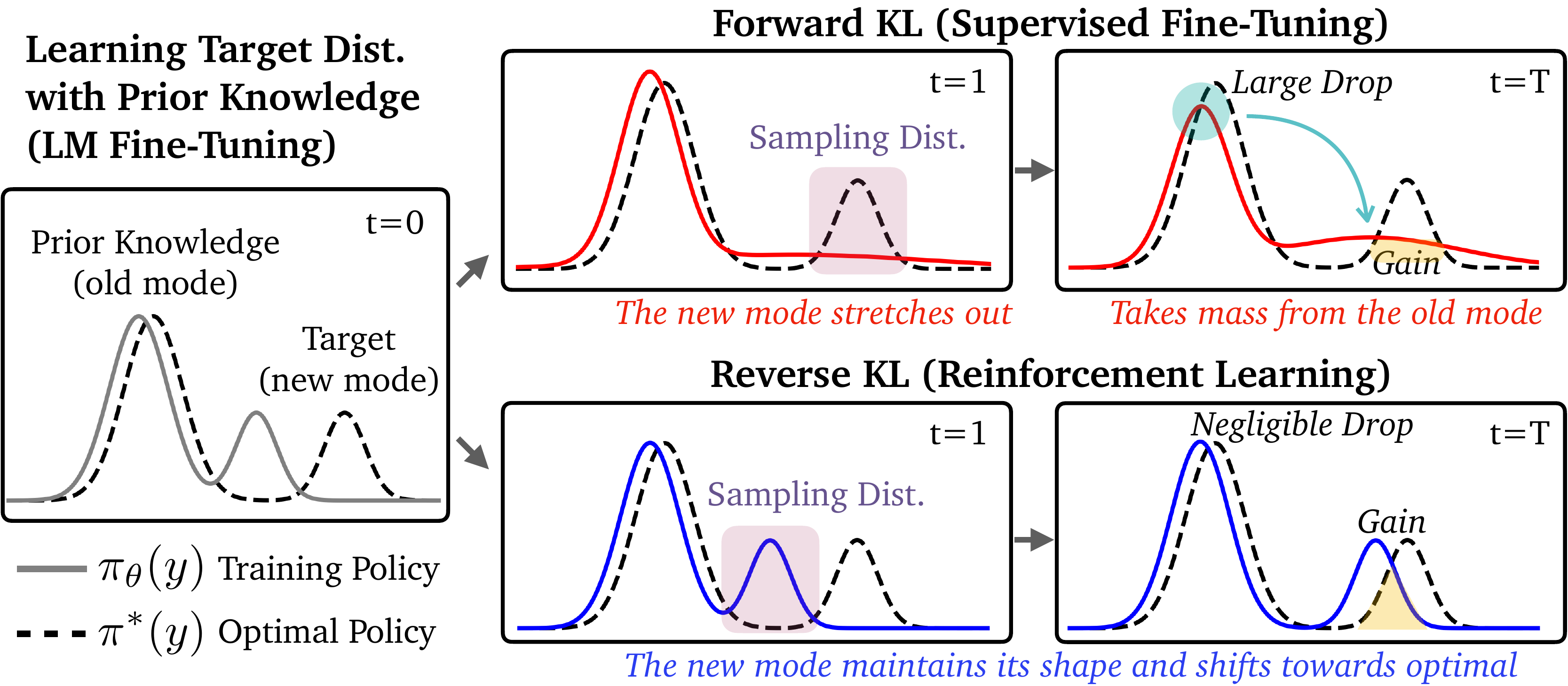}
    \caption{Forward KL (SFT) versus reverse KL (RL). SFT can take mass from the prior mode; RL tends to preserve it while amplifying the target mode. Adapted from \citet{chen2025retaining}.}
    \label{fig:sft-vs-rl}
\end{figure}

\textbf{Forgetting.} The standard objection to specialization is catastrophic forgetting. \citet{chen2025retaining} identify on-policy data as an important factor in the lower forgetting observed for RL post-training relative to SFT. Our dual-path design is consistent with that observation: SAO uses on-policy rollouts and clipped updates, while Evo-Harness does not update weights at all. We do not run a held-out retention test, so we treat retention as a design motivation rather than an empirical result in this paper; for the harness path, recovery is a file-level rollback rather than checkpoint restoration. Figure~\ref{fig:sft-vs-rl} summarizes the intuition: forward-KL-style SFT moves mass away from the old mode, while reverse-KL-style RL amplifies an existing mode.

\textbf{Contributions.}
\begin{itemize}[leftmargin=1.2em,itemsep=0.15em,topsep=0.2em]
    \item We formulate agent specialization as joint online adaptation of policy weights $\theta$ and harness workspace $\harness$ over a shared task-level rollout pattern, and state a retention constraint that this study does not yet enforce with a held-out gate.
    \item We engineer the weight loop as a real-time SAO RL system for 200k-token agent traces: exact capture and action-token-only training, turn-level credit with a task baseline, fused critic/actor objectives with prompt-prefix reuse, bounded chunking, and restart-based quantized vLLM serving.
    \item We instantiate the harness loop with one-shot Evo-Harness: each task is executed once per cycle, reflection is grounded in verifier and tool evidence, and the accepted result is an auditable workspace-file delta rather than a new candidate prompt population.
    \item We report completed SAO-only and Evo-only runs on six SWE-bench Verified instances, and audit solve rates, process rewards, token accounting, reasoning lengths, and implementation costs to characterize complementary mechanisms rather than declare a single winner.
    \item We describe workspace adapters and composition as deployment mechanisms, leaving base, combined, cross-domain, and retention-gate evaluations explicit future work rather than inferring them from the completed runs.
\end{itemize}

\section{Related Work}

\paragraph{RL for agents.} PPO \citep{schulman2017ppo} introduced the clipped surrogate objective that remains the default for stable policy updates. GRPO \citep{shao2024deepseekmath} removes the value network by normalizing rewards within a group of rollouts, reducing memory but creating a synchronization barrier: all rollouts for a prompt must finish before an update. SAO \citep{hou2026sao} removes that barrier with single-rollout asynchronous updates, token-level double-sided clipping (DIS), fast detached-feature value updates, and skip-observation GAE for multi-turn trajectories. \citet{abgoswam2024swegrpo} provide a practical account of GRPO on SWE-Bench that informs our infrastructure choices. Our contribution is not a new policy-gradient estimator: it is the serving, capture, credit-assignment, and streaming-training system that makes a single 200k-token tool-interacting rollout usable for real-time full-parameter updates.

\paragraph{Harness and tool use.} SWE-agent \citep{yang2024sweagent} showed that the agent-computer interface is a first-class design variable for software engineering. ReAct \citep{yao2023react} interleaves reasoning and acting; Toolformer \citep{schick2023toolformer} teaches models to call tools; Voyager \citep{wang2023voyager} builds an open-ended skill library. Reflexion \citep{shinn2023reflexion} and Self-Refine \citep{madaan2023selfrefine} use verbal feedback to improve behavior without gradients. LangChain \citep{langchain2023}, AutoGPT \citep{autogpt2023}, and Claude Code \citep{anthropic2024claude-code} package these ideas into modular harnesses. Our work differs in optimizing the harness itself rather than a single prompt or a fixed workflow.

\paragraph{Evolutionary and reflective optimization.} GEPA \citep{agrawal2025gepa} evolves prompts by reflecting on execution traces and maintaining a Pareto frontier; it reports larger gains than GRPO with far fewer rollouts on several reasoning tasks. OPRO \citep{yang2024opro} uses an LLM as an optimizer over prompt candidates, and EvoPrompt \citep{guo2024evoprompt} combines evolutionary operators with LLM-generated mutations. These methods treat the prompt string as the primary adaptation unit. Evo-Harness instead formulates online harness learning for a frozen solver: one-shot execution contexts are reflected into candidate lessons and compiled into persistent workspace-file edits \citep{wei2026evoharnesscontexttoharnessskillcompilation}. Our harness path adopts this mechanism so that adaptation is a structured file delta rather than a population of prompt variants, and so that a task is executed once rather than repeatedly sampled to score candidate prompts.

\paragraph{Adapters and composition.} LoRA \citep{hu2021lora} adapts weights with low-rank updates; AdapterFusion \citep{pfeiffer2021adapterfusion} composes task adapters without retraining the base model. Harness adapters follow the same deployment principle at the scaffolding level: small workspace deltas that could be loaded, unloaded, and composed. This is a design analogy, not a claim that workspace files implement low-rank parameter updates or learned adapter fusion.

\paragraph{Forgetting and retention.} EWC \citep{kirkpatrick2017catastrophic} penalizes changes to parameters important for old tasks. \citet{chen2025retaining} identify on-policy data as an important factor in RL's lower forgetting relative to SFT. Our harness path avoids weight-level forgetting by leaving $\theta$ unchanged, but it does not guarantee retention of prior harness behavior.

\paragraph{Benchmarks.} SWE-bench Verified \citep{jimenez2024swebenchlanguagemodelsresolve} measures repository-level issue resolution and is the only benchmark used in this study. Other agent benchmarks such as OfficeBench \citep{wang2024officebench}, Terminal-Bench \citep{merrill2026terminalbench}, and AgentBench \citep{liu2024agentbench} stress complementary capabilities; we leave cross-domain evaluation to future work.

\section{Problem Formulation and Method}

\subsection{Agent decomposition}
Let a task instance be $x\sim\mathcal{T}$. Conceptually, write the harness as
\begin{equation}
\harness=(P,\,T,\,S,\,M,\,\Phi),
\end{equation}
where $P$ is the system prompt and reasoning template, $T$ is the set of tool schemas, $S$ is the skill library, $M$ is the memory/context policy, and $\Phi$ is the planning/reflection procedure. In the implementation these fields are materialized as workspace files: a system prompt, tool registry, skills, exemplars, and a manifest; there is no first-class $M$ or $\Phi$ object. A rollout is
\begin{equation}
\tau=(a_1,o_1,\dots,a_K,o_K)\sim p(\tau\mid x,\policy,\harness),
\end{equation}
where actions $a_t$ are generated by the policy and observations $o_t$ come from the environment. The environment returns a scalar reward $R(x,\tau)$ at episode end, typically a test pass/fail or a task-completion score.

The specialization objective (Eq.~\ref{eq:objective}) is
\begin{equation}
\max_{\theta,\,\harness}\;\; J(\theta,\harness)
=\mathbb{E}_{x\sim\mathcal{T}}\,\mathbb{E}_{\tau\sim p(\cdot\mid x,\policy,\harness)}
\left[R(x,\tau)\right],
\label{eq:objective}
\end{equation}
subject to a retention constraint on a reference distribution $\mathcal{D}_{\old}$:
\begin{equation}
J_{\old}(\theta,\harness)\;\ge\;J_{\old}(\theta_0,\harness_0)-\epsilon.
\label{eq:retention}
\end{equation}
Equation~\ref{eq:retention} makes the trade-off explicit: a specialization method may improve the target domain, but it may not destroy prior capability. We state it as a design constraint rather than a measured result: the present study has no held-out retention gate. SAO uses on-policy data and clipped updates, while Evo-Harness leaves $\theta=\theta_0$ and makes its change auditable at the file level; neither observation establishes retention of prior behavior.

\subsection{Two outer loops over one inner loop}
The two specialization mechanisms operate on different time scales and consume rollouts produced by the same task-level interaction pattern, rather than the same run. The weight loop forms a full-parameter update
\begin{equation}
\theta_{t+1}=U_{\theta}(\theta_t,\tau_t,R_t),
\end{equation}
after each SAO update, while the harness loop accumulates execution contexts and performs a workspace revision
\begin{equation}
\harness_{i+1}=U_{\harness}(\harness_i,\{c_{i,j}\}_{j=1}^{n_i},F_i),
\end{equation}
after each batch. Here $c=(x,\tau,y,f)$ contains the task, trajectory, outcome, and verifier/tool feedback, and $F_i$ denotes the batch-level feedback aggregated for compilation. The weight loop changes the model's prior; the harness loop changes prompts, tools, skills, memory policy, and planning files in a workspace that can be snapshotted and rolled back. We study each loop independently and discuss their composition as a design goal. In the completed runs reported below, the two specialized artifacts use the same base model, task subset, and 400-turn budget, but their serving and network settings differ (Section~\ref{sec:setup}); we do not evaluate their post-hoc composition, and online interleaving of the two updates is left to future work.

\subsection{Inner loop: environment interaction}
Both loops use the same terminal-agent interaction pattern. Tasks run with up to 400 turns and a 200k-token model window; SAO caps each policy response at 24,576 tokens, while the Evo run records \texttt{max\_tokens=auto} and gateway requests used 200,000. Both enable progress-stall detection at 200 steps; the Evo timeout observed in this study is 3,000 seconds, while SAO records \texttt{timeout\_seconds=auto} and has no timeout. At each turn the harness assembles a context from the transcript, the active tools, and loaded skills; the policy emits an action; the environment executes it and returns an observation. The episode ends on an explicit submit action, budget exhaustion, a timeout, or an unrecoverable error. This separation is important: the inner loop is the \emph{data source} for both outer loops, even though serving and network settings are not identical.

\subsection{Outer loop A: online harness learning with Evo-Harness}
The harness loop keeps $\theta$ frozen when run alone and optimizes the whole harness workspace $\harness$ rather than an isolated prompt string or skill library. The specialization unit is an auditable workspace-file delta: any prompt, tool description, skill, exemplar, planning file, or manifest entry can be revised. The current implementation prefers editing an existing file and creates a new skill only when no existing file can carry the lesson. A workspace can be materialized as a deployment adapter $\adapter_{\dom}=\delta(\harness_{\dom})$ without changing the base checkpoint.

Let $\mathcal{X}$ be the training tasks and $\harness$ the seed workspace. Each task $x$ is executed exactly once under the current workspace:
\begin{equation}
c_x=(x,\tau_x,y_x,f_x,p_x)=\textsc{Run}(\policy,\harness,x),
\end{equation}
where $\tau_x$ is the trajectory, $y_x$ the outcome, $f_x$ verifier or tool feedback, and $p_x$ the exact capture of model turns. A single execution may contain many model and tool calls; one-shot means that no original task is replayed to score candidate workspaces. After all tasks in a cycle, the evidence index $\mathcal{E}$ exposes rollout records, captures, task files, and Docker listings to the reflection model, which returns one lesson and complete-file changes:
\begin{equation}
(\ell,\Delta)=\textsc{ReflectWorkspace}(\mathcal{E},\harness).
\end{equation}
The compiler writes $\Delta$ into a copy of the workspace, validates it, and the accepted final edit replaces the seed workspace:
\begin{equation}
\harness'=\textsc{ReplaceWorkspace}\bigl(\textsc{Validate}(\textsc{ApplyEdits}(\harness,\Delta))\bigr).
\end{equation}
If $\Delta$ is empty, the lesson is appended to \texttt{prompts/system.md}; deletions of \texttt{manifest.yaml} or \texttt{prompts/system.md} are rejected. There is no held-out before/after gate, \texttt{best} selection, or adapter produced by training. Algorithm~\ref{alg:evo} summarizes this loop.

\begin{algorithm}[t]
\caption{One-shot evidence-to-workspace evolution (Evo-Harness)}
\label{alg:evo}
\begin{algorithmic}[1]
\Require training tasks $\mathcal{X}$, frozen solver $\policy$, seed workspace $\harness_0$, reflection model $q$
\Ensure evolved workspace $\harness^\star$ and audit trail
\State $\harness \gets \textsc{ValidateWorkspace}(\harness_0)$; $\mathcal{C}\gets\emptyset$
\ForAll{$x\in\mathcal{X}$}
    \State $\harness_x \gets \textsc{MaterializeWorkspace}(\harness,x)$
    \State $(\tau,y,f,p) \gets \textsc{RunOnce}(\policy,\harness_x,x)$
    \State $\mathcal{C}\gets\mathcal{C}\cup\{(x,\tau,y,f,p)\}$
\EndFor
\State $\mathcal{E}\gets\textsc{BuildEvidenceIndex}(\mathcal{C})$
\State $(\ell,\Delta)\gets\textsc{ReflectWorkspace}(q,\mathcal{E},\harness)$
    \Comment{study successes and failures; prefer edits to existing files}
\If{$\Delta=\emptyset$}
    \State $\Delta\gets\textsc{AppendLessonToSystemPrompt}(\ell)$
\EndIf
\State $\harness'\gets\textsc{ApplyEdits}(\harness,\Delta)$
\State $\textsc{ValidateWorkspace}(\harness')$ \Comment{path safety; manifest and system prompt survive}
\State $\harness^\star\gets\textsc{ReplaceWorkspace}(\harness,\harness')$
\State \Return $\harness^\star$, $\ell$, $\Delta$, $\mathcal{E}$
\end{algorithmic}
\end{algorithm}

The completed run records one accepted workspace-evolution cycle over all six training tasks: four tasks pass, one fails verification, and one times out. The artifact log contains three reflection/compilation attempts on the same rollout evidence; the final accepted workspace contains one auditable edit to \texttt{prompts/system.md} and no new skills.

\subsection{Outer loop B: weight updates with SAO RL}
The weight loop performs full-parameter policy updates with SAO \citep{hou2026sao}. For long-horizon agentic tasks, SAO replaces GRPO's group-wise sampling with one rollout per task, so no group synchronization barrier is needed. Let $\ell_t^{\old}$ and $\ell_t^{\new}$ be the captured and recomputed log-probabilities of action token $t$, and let $m_t$ be the assistant-action mask. The token-level ratio and double-sided clipping (DIS) are
\begin{equation}
\rho_t=\exp\!\bigl(\operatorname{clip}(\ell_t^{\new}-\ell_t^{\old},-M_{\mathrm{KL}},M_{\mathrm{KL}})\bigr),\qquad
\tilde\rho_t=
\begin{cases}
\rho_t, & 1-\epsilon_{\ell}<\rho_t<1+\epsilon_h,\\
0, & \text{otherwise}.
\end{cases}
\end{equation}
The strict interval means that tokens outside the trust region receive zero actor weight; they still remain in the mask-normalization denominator. The clipping bounds are fixed run-level hyperparameters, not per-task fitted values.

Skip-observation GAE runs backward over active action tokens only. With process reward $r_t$ placed on the last masked action token of each turn and terminal reward added on the final turn,
\begin{equation}
\delta_t=r_t+\gamma V_{t+1}-V_t,\qquad
G_t^{A}=\delta_t+\gamma\lambda_t^{A}G_{t+1}^{A},\qquad
G_t^{V}=\delta_t+\gamma\lambda_V G_{t+1}^{V},
\end{equation}
where masked observation tokens are skipped without changing the carried value. The actor uses length-adaptive $\lambda_t^{A}=\operatorname{clip}(1-1/(\alpha q_t),0,1)$ with $q_t$ the number of active tokens from $t$ to the end; the critic uses fixed $\lambda_V$. The policy loss is normalized by the number of active masked tokens, and the scalar value head is updated twice from detached backbone features before one full-model actor step. Algorithm~\ref{alg:sao} gives the implemented schedule.

\begin{algorithm}[t]
\caption{Single-rollout SAO RL update}
\label{alg:sao}
\begin{algorithmic}[1]
\Require tasks $x_1,\dots,x_U$, initial policy $\theta_0$, fixed harness $\harness$, updates $U$, $\epsilon_\ell$, $\epsilon_h$, $M_{\mathrm{KL}}$, $\gamma$, $\lambda_V$, $\alpha$, learning rates $\eta_\pi,\eta_V$, critic steps $C$
\Ensure trained checkpoints $\theta_1,\dots,\theta_U$
\For{$u=1$ to $U$}
    \State serve $\theta_{u-1}$; collect one rollout $\tau_u$ for $x_u$ under $\harness$
    \State capture exact $(p_i,a_i,\ell_i^{\old},m_i)$ for each model turn
    \State bucket process events into turn-level rewards and place $r_t$ on the last masked action token
    \State add terminal reward $R_u-b(x_u)$ to the final turn; $b(x_u)=0$ for single-sample tasks
    \For{each active action token $t$}
        \State $q_t\gets$ number of active tokens from $t$ to the end
        \State $\lambda_t^A\gets\operatorname{clip}(1-1/(\alpha q_t),0,1)$
    \EndFor
    \State run one fused prefix-exact forward/backward to obtain $\ell_t^{\new}$, $V_t$, and detached critic features
    \For{$t$ from the last active token to the first}
        \State skip masked tokens; $\delta_t\gets r_t+\gamma V_{t+1}-V_t$
        \State $G_t^A\gets\delta_t+\gamma\lambda_t^A G_{t+1}^A$; $G_t^V\gets\delta_t+\gamma\lambda_V G_{t+1}^V$; $Y_t\gets V_t+G_t^V$
    \EndFor
    \State $\rho_t\gets\exp(\operatorname{clip}(\ell_t^{\new}-\ell_t^{\old},-M_{\mathrm{KL}},M_{\mathrm{KL}}))$
    \State keep token $t$ only if $1-\epsilon_\ell<\rho_t<1+\epsilon_h$; otherwise set its detached weight $w_t=0$
    \State $\mathcal{L}_\pi\gets-\sum_t m_t w_t G_t^A\ell_t^{\new}/\sum_t m_t$
    \For{$c=1$ to $C$}
        \State $\mathcal{L}_V\gets\sum_t m_t(V_t-Y_t)^2/\sum_t m_t$ on detached features
        \State take one value-head step with $\eta_V$
    \EndFor
    \State take one full-model actor step with $\eta_\pi$; save and verify $\theta_u$
\EndFor
\State \Return $\theta_1,\dots,\theta_U$
\end{algorithmic}
\end{algorithm}

The completed run uses six updates, strict clipping bounds $\epsilon_\ell=0.8$ and $\epsilon_h=3.0$, two critic steps per actor step, and exact capture for all 329 training responses.

\subsection{System co-design for real-time long-horizon rollouts}
\label{sec:system}
The SAO instantiation is a systems design around one constraint: a single tool-interacting episode can approach 200k tokens, yet the next rollout must be served by the updated checkpoint. We therefore separate serving, exact capture, and training instead of treating the estimator as an isolated equation.

\textbf{Rollout/training boundary.} Rollouts are served by vLLM in a separate process because online RL needs same-format HF parameter names and dtypes, token-level logprobs and token ids, revisioned old logprobs, and high rollout throughput. The capture freezes prompt ids, completion ids, and old logprobs; training recomputes new logprobs and gradients and does not reuse inference hidden states or KV caches. The training step runs in a child process; after it exits, the orchestrator stops and restarts the vLLM process group to load the new checkpoint. Explicit cache cleanup calls are safeguards inside this isolation design; we do not claim a measured zero-residue invariant.

\textbf{Credit assignment and action tokens.} Each task is sampled once per update. Process rewards are placed on the last assistant action token of each turn; the terminal reward is added on the final turn, with a task-level baseline only when the same task has multiple samples. In the completed single-sample run this baseline is zero. We use turn-level SAO advantages with skip-observation GAE and length-adaptive actor $\lambda$, not group mean/std, \texttt{pass@k} margins, or group-relative length anchors. Statistics such as final-context length are computed from final-trace snapshots, whereas gradients use all assistant completions in the full trace with \texttt{response\_mask=1}; cumulative prompts are state and model completions are actions.

\textbf{Exact streaming and fused objectives.} Traces are not truncated: when the exact-chunk contract is violated, the implementation fails rather than silently shortening a rollout. Stage~1 streams detached prefix K/V with CPU offload, stage~2 recomputes the current chunk with backward, and \texttt{prepare} skips the backbone forward so event reordering alone does not save VRAM. The fused actor objective produces critic features, values/GAE, and the actor loss in the same chunked pass. When every turn prompt is a strict prefix of the final prompt, the stage~1 prefix cache is computed once under \texttt{no\_grad} and reused with detached K/V; each completion still runs stage~1 and stage~2, and divergence falls back to the conservative per-turn path.

\begin{table}[t]
\centering
\caption{Exact-streaming implementation decisions. Forward-call counts use a fake CPU model only for call counting and numerical agreement, not wall-clock speed; the memory figure is a single measured RTX 4060 Laptop setting.}
\label{tab:streaming}
\footnotesize
\setlength{\tabcolsep}{4pt}
\begin{tabular}{p{0.24\linewidth}p{0.62\linewidth}}
\toprule
Decision & Implementation and evidence \\
\midrule
Fused objective & A chunked actor pass yields critic features, GAE, and actor loss instead of a separate critic-feature forward; calls fall from 543 to 362 on the last turn and 17{,}073 to 11{,}382 over 56 turns ($-33\%$). \\
Prepare ordering & Advancing \texttt{prepare-start} does not reduce VRAM, because exact-streaming prepare skips the backbone forward; the gain comes from fusion. \\
Prompt-prefix reuse & A detached \texttt{no\_grad} stage~1 prefix cache is shared when prompts are strict prefixes; completion stage~1 and stage~2 still run per turn, with fallback on divergence. Calls fall to 184 and 474 ($-66\%$ and $-97\%$ versus the original path). \\
Chunk bounds & On an RTX 4060 Laptop 8GB, \texttt{chunk\_rows=512} peaks at 6.44\,GiB; 1024 and 2048 OOM in the tested setting, so larger chunks are not assumed. \\
Numerical checks & Critic features, advantages, returns, and the objective agree with the unfused path under the implementation tests. \\
\bottomrule
\end{tabular}
\end{table}

\subsection{Workspace specialization and adapter composition}
Training replaces the seed workspace with an evolved workspace $\harness_{\dom}$; for deployment, the workspace can be materialized as an adapter $\adapter_{\dom}=\delta(\harness_{\dom})$. This adapter utility is not a trained artifact of the completed run. An adapter modifies a subset of workspace fields, and composition is defined when the modified namespaces are disjoint:
\begin{equation}
(\harness\oplus\adapter_1)\oplus\adapter_2
=\harness\oplus(\adapter_1\oplus\adapter_2).
\end{equation}
For overlapping fields the current implementation uses right-biased override: the later adapter wins. This mirrors the non-destructive deployment principle of AdapterFusion \citep{pfeiffer2021adapterfusion} at the file level, but it is neither a parameter-space update nor a learned fusion module. As a design illustration, a SWE workspace could own the code-editing prompt section and test/debug skills, an Office workspace could own document/UI tools, and a CLI workspace could own shell-specific tool descriptions; none of these cross-domain variants is evaluated here. The policy $\policy$ is shared, and training itself does not select a \texttt{best} adapter.

\section{Experimental Setup}
\label{sec:setup}

\subsection{Base agent: Codev}
Codev \citep{chen2026codev} is a Claude Code-style terminal agent with a modular harness. The SWE seed workspace used in this study contains a planning/reasoning system prompt, an eight-entry tool registry (\texttt{Bash}, \texttt{Edit}, \texttt{Glob}, \texttt{Grep}, \texttt{Read}, \texttt{Skill}, \texttt{TodoWrite}, \texttt{Write}), five skills, four exemplars, and context-management configuration with automatic compaction. Codev exposes a TUI and a headless mode; all completed runs use headless execution.

\begin{figure}[t]
    \centering
    \includegraphics[width=0.82\linewidth]{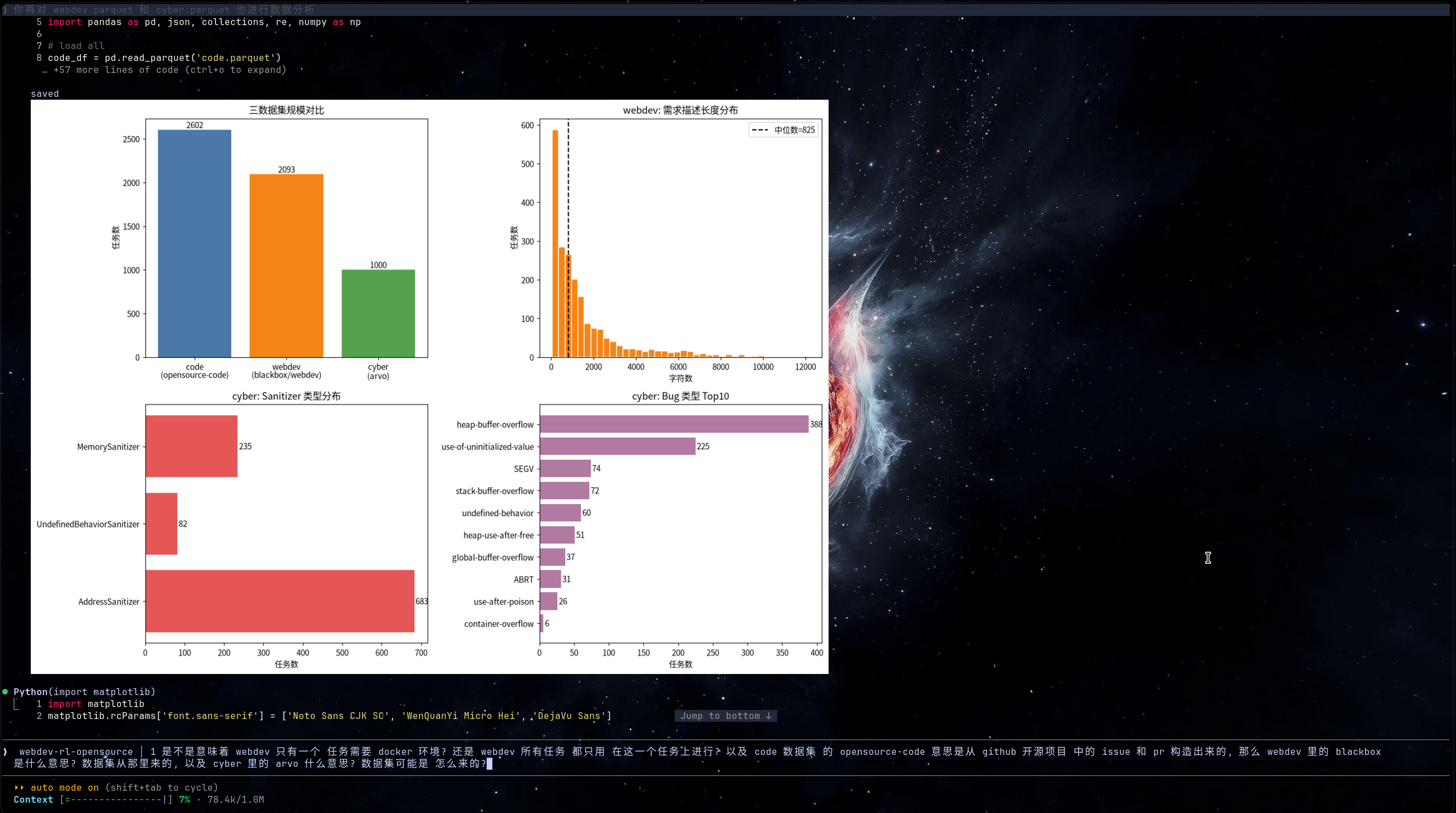}
    \caption{Illustrative Codev session for exploratory data analysis. The agent writes and runs plotting code, inspects the resulting figures, and iterates on its analysis. This illustrates tool-use and context-management behavior that harness design can shape; it is not a quantitative evaluation result.}
    \label{fig:codev-tui}
\end{figure}

\subsection{Benchmark and task subset}
We use SWE-bench Verified \citep{jimenez2024swebenchlanguagemodelsresolve} only. The completed runs cover six repository-level instances (Table~\ref{tab:benchmarks}); all reported solve rates are on these six training instances, not on the full benchmark. Each instance uses the task's verifier outcome as a binary reward.

\begin{table}[t]
\centering
\caption{SWE-bench Verified subset used in the completed runs.}
\label{tab:benchmarks}
\small
\begin{tabular}{llcc}
\toprule
Benchmark & Domain & Instances used & Metric \\
\midrule
SWE-bench Verified & Software engineering & 6 & Binary reward \\
\bottomrule
\end{tabular}
\end{table}

\subsection{Configurations}
We compare four configurations:
\begin{enumerate}[leftmargin=1.4em,itemsep=0.1em,topsep=0.2em]
    \item \textbf{Base:} original Codev $(\policy,\harness)$; no completed run artifact is available for this configuration.
    \item \textbf{SAO-only:} full-parameter SAO-trained $\policy^*$ with the original $\harness$.
    \item \textbf{Evo-only:} original $\policy$ with the evolved workspace $\harness_{\dom}$.
    \item \textbf{Combined:} SAO-trained $\policy^*$ with the evolved workspace $\harness_{\dom}$; this post-hoc composition is defined but not evaluated here.
\end{enumerate}
The completed SAO-only and Evo-only runs use the same base-model path, task subset, 400-turn budget, and 200k context window; only the specialized object differs in principle. Docker images are task-derived rather than fixed in the run configs, and response caps differ: SAO records 24,576 tokens while Evo records \texttt{max\_tokens=auto}. The base and combined configurations are defined for completeness but have no completed run artifacts in this study.

\subsection{Serving stack and precision}
The rollout server is vLLM rather than a converted runtime format because online RL needs weights and captures to share parameter names and dtypes, token ids and token-level logprobs, and a serving process with sufficient throughput; old logprobs are tied to a recorded weight version. The completed path uses the BF16 source checkpoint as the canonical training artifact but serves with an FP8 target, \texttt{fp8\_e4m3} KV cache, and DSpark draft speculative decoding (seven speculative tokens, greedy draft sampling) at a 200k model window. MXFP4 is supported at the configuration and test level, but the completed run does not evaluate an MXFP4 end-to-end serving path, and full BF16/FP8/MXFP4 logprob and throughput curves are not reported. DSpark changes the server generation path, not the OpenAI-compatible wire schema, and quantized serving uses restart rather than packed-weight hot patching because FP8/MXFP4 layouts do not preserve the IPC update contract. A local Codev comparison reported more stable vLLM rollouts and lower resident memory than the prior llama.cpp server, but this is a systems observation rather than a controlled precision--quality ablation. Together, these choices make precision a serving-system decision rather than a change to the training objective.

\subsection{Compute budget}
The retained cost artifacts are windowed rather than a single end-to-end ledger. SAO's \texttt{training-cost.json} lists six checkpoints and records 0.567 GPU-hours and 0.57 hours of wall time for the final resumed training window; aggregate GPU time across all resumed sessions is not available. Evo-Harness performs no policy weight training, and its accepted reflection/compilation window records 19.8 seconds and 0.0055 GPU-hours. Its six completed rollouts sum to 2.51 hours and 20.46M API tokens in \texttt{cycles.jsonl}, while the run-level budget fields separately record 62.65M rollout tokens and 0.304M evolver tokens; these accountings cannot be reconciled from the retained artifacts. We therefore report both all-turn API tokens and final-trace snapshots, and label their sources explicitly. The combined configuration is a proposed composition, not an evaluated or separately trained run.

\begin{table}[t]
\centering
\caption{Artifact-windowed compute for the completed runs. SAO GPU/wall time covers the retained final training window and lists six checkpoints; Evo wall time combines completed rollout latency with the accepted 19.8s reflection window. API tokens count every recorded turn in the indicated artifact; final trace tokens are last-turn snapshots and are medians over six tasks, with the Evo timeout counted as zero. Combined is defined but not evaluated.}
\label{tab:compute}
\footnotesize
\setlength{\tabcolsep}{4pt}
\begin{tabular}{lcccc}
\toprule
Config & Weight-train GPU h & Wall time & API tokens & Final trace tokens \\
 & & & (completed rollouts) & (median/task) \\
\midrule
Base & 0 & --- & --- & --- \\
SAO-only & 0.567 (window) & 0.57h monitored & 16.8M & 92{,}463 \\
Evo-only & 0 (no policy training) & 2.51h rollout; 19.8s evolve & 20.5M & 74{,}211 \\
Combined & --- & --- & --- & --- \\
\bottomrule
\end{tabular}
\end{table}

\subsection{SAO hyperparameters}
The completed SAO run starts from \texttt{spark-x2.6-1.7b-fp8} (\texttt{Spark2\_5ForCausalLM}), served as \texttt{spark-sao} with \texttt{tool\_call\_parser=spark25}; the \texttt{qwen3} reasoning parser only delimits model output and does not indicate a Qwen checkpoint. The run uses six updates, $\epsilon_{\ell}=0.8$, $\epsilon_{h}=3.0$, learning rate $1\times10^{-6}$, $\gamma=1.0$, $\lambda=1.0$, length-adaptive policy GAE with $\lambda_{\alpha}=0.5$, $\lambda_{\text{critic}}=1$, and two critic steps per actor step. Rollouts use \texttt{max\_tokens}=24576 and exact trace-token capture; all 329 training responses preserve prompt ids, completion ids, and logprobs.

\subsection{Evo-Harness settings}
The completed Evo-Harness run records one accepted workspace-evolution cycle over all six training tasks; the recorded batch size does not create additional accepted cycles. Each task is executed once with standard verifier feedback. Four tasks pass; one fails verification and one hits a 3,000-second execution timeout, which we retain as a real infrastructure failure rather than relabeling it as a model error. The artifacts record three reflection/compilation attempts on the same rollout evidence: an empty change list, a rejected skill-file proposal, and the accepted edit. Reflection sees the completed episodes and the full seed workspace, prioritizes failures and negative feedback, and returns one lesson plus concrete file changes. The final accepted workspace contains one edit, to \texttt{prompts/system.md}, and no new skills. Edits may add, revise, or delete files under \texttt{recipes/harness/swe\_evo/seed\_workspace}; empty change lists are converted into an auditable append to \texttt{prompts/system.md}. After validation, the accepted workspace replaces the seed workspace. There is no candidate task re-execution, held-out before/after gate, selected \texttt{best} adapter, or per-cycle checkpoint.

\subsection{Evidence sources and accounting}
Table~\ref{tab:evidence} records which artifacts support the empirical claims. We keep differences between ledgers visible rather than forcing them into one number: in particular, all-turn API tokens, final-turn snapshots, and run-level budgets answer different questions, and the Evo timeout has captured gateway responses but no completed episode record.

\begin{table}[t]
\centering
\caption{Artifact sources for the empirical audit. ``Completed rollout'' accounting excludes captured responses that never formed a completed episode.}
\label{tab:evidence}
\footnotesize
\setlength{\tabcolsep}{4pt}
\begin{tabular}{p{0.22\linewidth}p{0.30\linewidth}p{0.40\linewidth}}
\toprule
Claim family & Primary artifact & Accounting note \\
\midrule
Terminal outcomes & \texttt{state/completed.jsonl}, \texttt{state.json}, per-update \texttt{results.jsonl} & Binary verifier outcome; Evo timeout has no completed episode. \\
Process rewards & \texttt{process\_rewards} in completed records; \texttt{events.jsonl} for SAO & Per-task or per-update means; the two grand means both round to 0.414. \\
Tokens & \texttt{cycles.jsonl}, \texttt{state.json}, \texttt{training-cost.json} & Completed-rollout API totals and final-trace medians differ from run-level budget fields. \\
Reasoning lengths & \texttt{captures/*.jsonl}, streamed reasoning deltas from HTTP 200 responses & Evo includes 101 timeout-capture responses; SAO covers 329 exact training turns. \\
\bottomrule
\end{tabular}
\end{table}

\section{Results}

\subsection{Observed outcomes on the six instances}
Table~\ref{tab:main-scores} reports only outcomes supported by completed run artifacts. SAO-only resolves all six rollouts; Evo-only resolves four, with one verifier failure and one timeout. We do not have completed base or combined evaluations, so we do not report scores for them. The goal is to compare solve, process, and cost profiles, not to claim benchmark superiority.

\begin{table}[t]
\centering
\caption{Observed outcomes on six SWE-bench Verified training instances. ``Not evaluated'' means no completed run artifact is available; no score is inferred.}
\label{tab:main-scores}
\small
\begin{tabular}{lccc}
\toprule
Config & Specialized object & Resolved & Failure mode \\
\midrule
Base & none & --- & not evaluated \\
SAO-only & $\policy^*$ & 6/6 & none \\
Evo-only & $\harness_{\dom}$ & 4/6 & 1 verifier failure; 1 timeout \\
Combined & $\policy^*+\harness_{\dom}$ & --- & not evaluated \\
\bottomrule
\end{tabular}
\end{table}

These outcomes are consistent with the process and cost audits below: SAO has the higher terminal solve count on these six instances, while Evo-Harness has no weight-training GPU cost and exposes a different process trajectory. Because the base and combined configurations are not evaluated, we make no claim about relative gain, sub-additivity, or domain-specific bottlenecks.

\subsection{Compute efficiency}
The two loops spend their budgets differently, but the retained artifacts do not support a single end-to-end efficiency ranking (Table~\ref{tab:compute}). SAO's retained cost window records 0.567 GPU-hours for the final resumed training session and lists six checkpoints; aggregate GPU time across all resumed sessions is unavailable. Evo-Harness performs no policy weight training. Its six completed rollouts sum to 2.51 hours and 20.46M API tokens, and the accepted reflection window adds 19.8 seconds and 0.0055 GPU-hours; a separate run-level budget reports 62.65M rollout tokens, which we cannot reconcile with the completed-rollout total. Evo-Harness executes each task once and does not re-run it for candidate selection, so the completed-rollout API total includes execution evidence and workspace context across every tool turn rather than repeated candidate validation. At the final-trace accounting, the Evo median is 74{,}211 tokens per task versus 92{,}463 for SAO; the Evo median counts the timed-out task as zero, while excluding it gives 81{,}129. The timed-out task has no completed episode or final-trace snapshot, although the gateway capture retains 101 model responses. These accountings are not interchangeable. The practical implication is not that one loop replaces the other: a team without a training cluster can obtain meaningful specialization from workspace edits, while a team with both budgets has a plausible composition to evaluate next.

\subsection{Real-time rollout implementation audit}
The engineering cost of exact long-horizon SAO is not captured by GPU-hours alone. Table~\ref{tab:streaming} shows why the implementation needed more than an off-the-shelf trainer: fusion removes a separate critic-feature pass, prefix reuse avoids repeating the growing prompt prefix, and bounded chunks keep an 8GB device within memory. These figures are call-count and single-device evidence, not a wall-clock speedup claim; the numerical tests only show agreement with the unfused path. The restart-based quantized serving path is likewise an enabling systems choice, not a reported precision-quality result. The harness loop has a different profile: it pays no policy-training GPU cost and avoids candidate re-execution by compiling grounded evidence into one workspace delta. These costs and artifacts are complementary operating points rather than components of a single ranking.

\subsection{Loop-specific diagnostics}
\begin{figure}[t]
    \centering
    \includegraphics[width=0.95\linewidth]{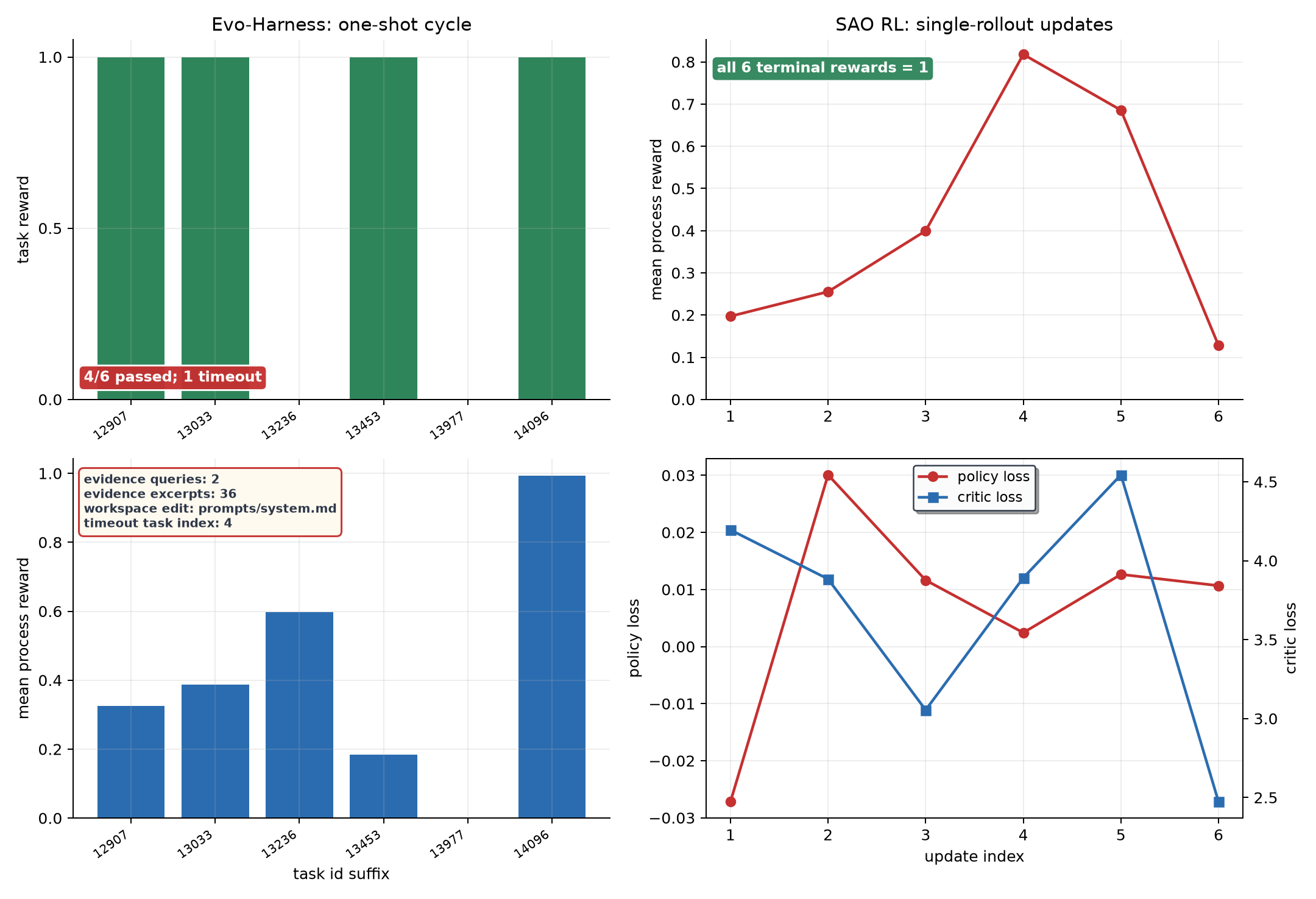}
    \caption{Loop-specific diagnostics. Left: Evo-Harness's accepted one-shot cycle over six tasks, with task reward, per-task process reward, evidence counts, and the final workspace edit after three reflection attempts. Right: SAO RL's six single-rollout updates, with process reward and actor/critic losses. The two sides use different x-axes and are not a paired comparison.}
    \label{fig:loop-profiles}
\end{figure}

Figure~\ref{fig:loop-profiles} reports each loop on its own mechanism rather than aligning the two runs by rollout index, because they visit tasks in different orders. The left panels describe Evo-Harness's accepted one-shot cycle: six task rewards, with the verifier failure and timeout marked, and per-task mean process reward together with the evidence and edit summary. The reflection stage issues two evidence queries, reads 36 excerpts, and attempts three workspace edits; the final accepted artifact edits \texttt{prompts/system.md} and creates no new skill. Its grand mean process reward is 0.4145. The right panels describe SAO RL's six full-parameter updates: process reward and actor/critic losses by update index. All six terminal rewards are 1, but the process signal is non-monotone (0.20, 0.26, 0.40, 0.82, 0.69, 0.13) with a grand mean of 0.4139. These views characterize each loop's own dynamics instead of reducing them to a shared score curve.

\subsection{Training cost and reasoning length}
\begin{figure}[t]
    \centering
    \includegraphics[width=0.95\linewidth]{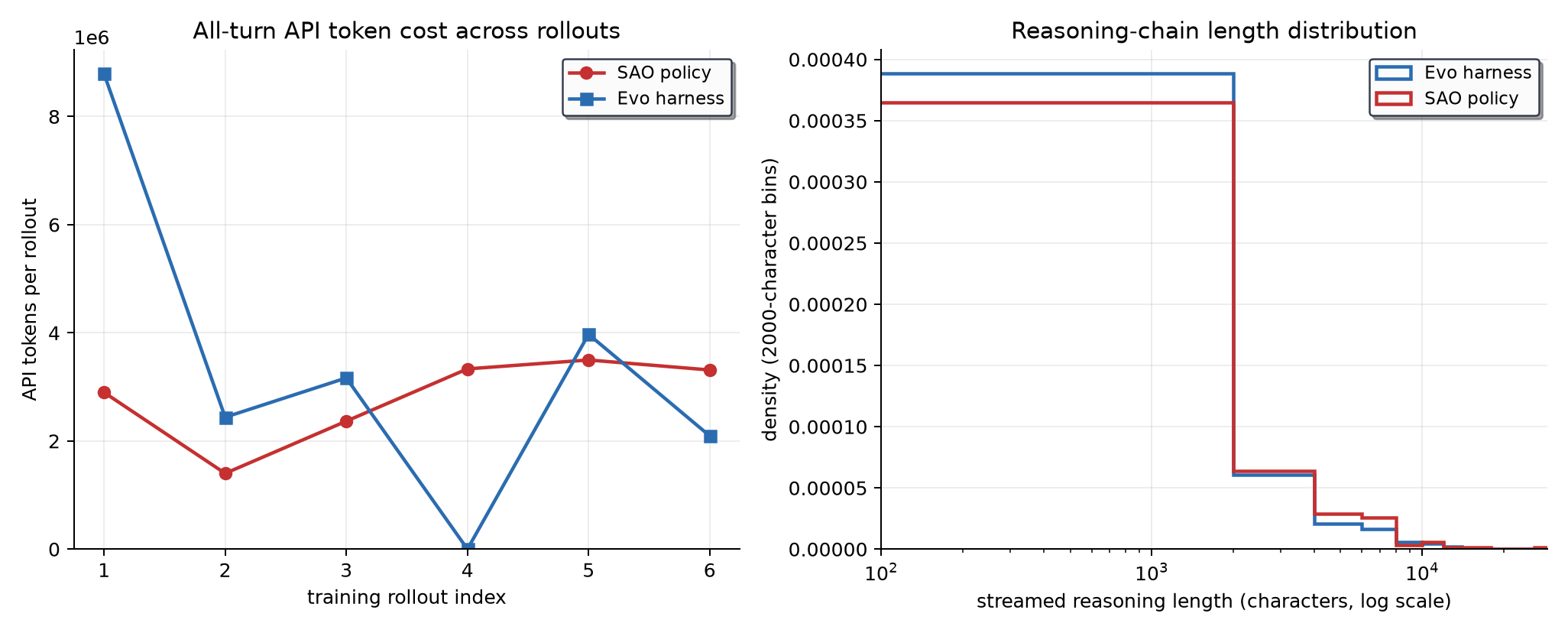}
    \caption{Left: all-turn API tokens per completed rollout; the Evo run includes one zero-token timeout. Right: streamed reasoning-length distributions in 2{,}000-character bins over HTTP 200 capture responses (Evo includes 101 responses from the timed-out task). Medians are 342 and 506 characters.}
    \label{fig:length}
\end{figure}

Figure~\ref{fig:length} pairs completed-rollout API token usage with the distribution of streamed reasoning-chain lengths in 2{,}000-character bins. The left panel uses all-turn accounting, so the token trajectories describe operating cost rather than a controlled efficiency ablation; the final-trace medians are reported separately in Table~\ref{tab:compute}. Reasoning lengths are not uniformly compressed: the medians are 342 characters for \texttt{spark-dspark} (452 HTTP 200 capture responses, including 101 from the timed-out task) and 506 characters for \texttt{spark-sao} (329 exact training turns).

\subsection{Cross-domain evaluation}
We do not train or evaluate adapters for OfficeBench or Terminal-Bench, so this study provides no evidence for cross-domain isolation or forgetting. Adapter composition remains a design principle to evaluate in future work.

\subsection{Qualitative observations}
\begin{figure}[t]
    \centering
    \includegraphics[width=0.9\linewidth]{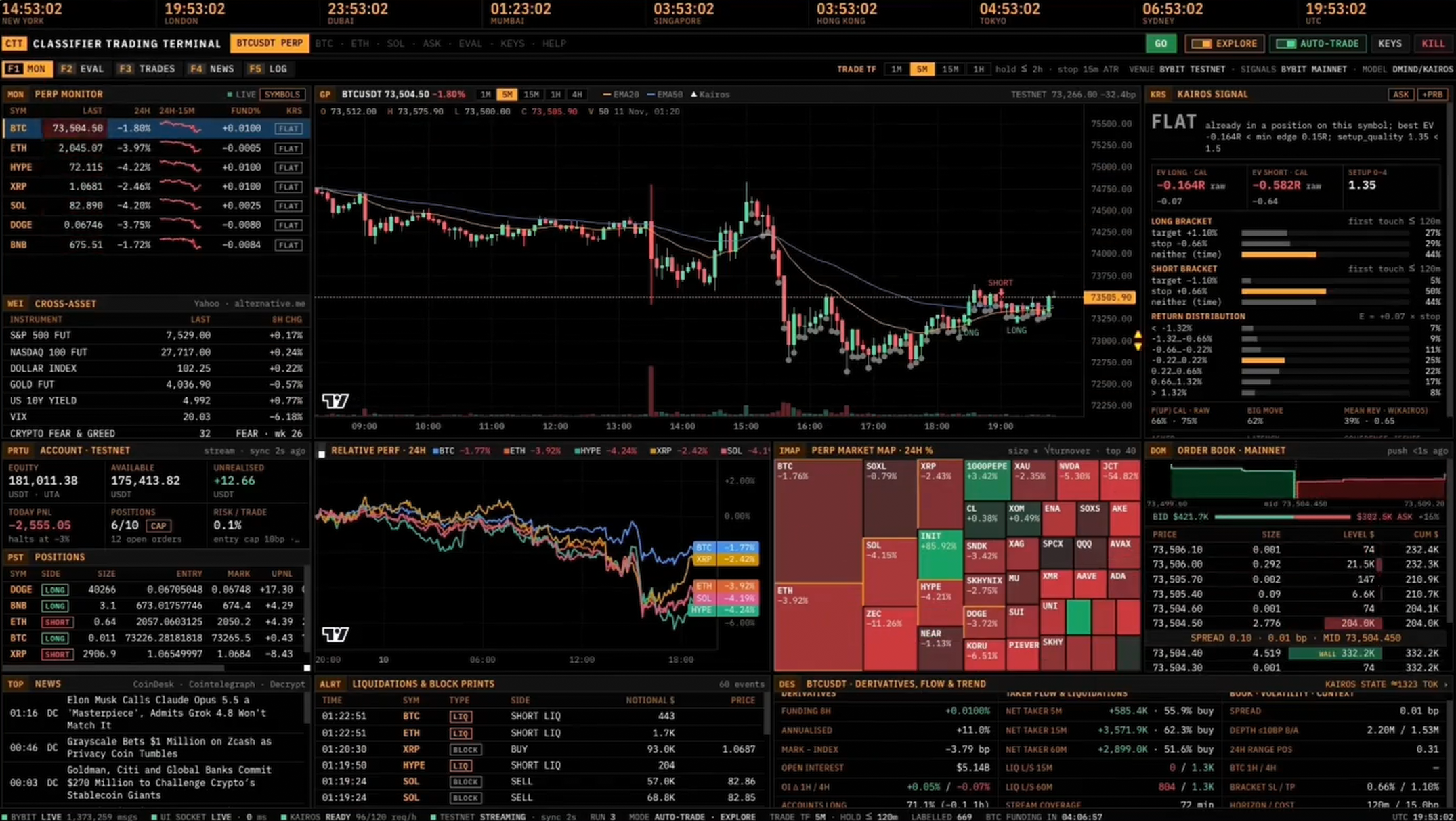}
    \caption{Illustrative domain-specific financial trading harness. The workspace contributes market-specific tools, terminology, and a risk-aware decision procedure; the underlying policy is unchanged. This is a design illustration, not an evaluation result.}
    \label{fig:domain-harnesses}
\end{figure}

Two qualitative illustrations show how workspace design can change behavior. In the Codev data-analysis session (Figure~\ref{fig:codev-tui}), the agent writes plotting code, runs it, inspects the generated figures, and revises the analysis. The useful harness behavior is a change in which existing policy capabilities are elicited: a sequence of tool calls and context summaries that keeps the analysis on track. In the illustrated financial-trading harness (Figure~\ref{fig:domain-harnesses}), domain-specific tools, risk vocabulary, and a decision procedure would be unnatural to bake into a coding prompt. These illustrations suggest that much of what looks like ``domain expertise'' in an agent may be harness design: which tools are visible, which examples are in context, and which planning template is active.

\section{Discussion}

\subsection{Why the two loops are complementary}
The loops are not interchangeable because they modify different objects. The weight loop changes the policy's conditional distribution and may be preferable when the target behavior is genuinely absent or when low-level fluency is the bottleneck; our SAO instantiation performs full-parameter on-policy updates. The harness loop changes the evidence, tools, and procedures supplied to the policy and may be preferable when behavior is latent but poorly elicited, or when interpretability and reversibility matter. The observed 6/6 versus 4/6 solve counts are descriptive, not a method ranking: the runs differ in serving and response caps, and neither base nor combined configurations were evaluated. Table~\ref{tab:decision} therefore summarizes affordances and observed evidence rather than assigning winners. The central claim is that a trained policy and an evolved workspace are jointly useful artifacts: one embeds behavior in weights, the other makes behavior easier to elicit, audit, and roll back.

\begin{table}[t]
\centering
\caption{Design affordances and observed evidence for the two completed loops. Solve counts are descriptive observations on six training instances, not a controlled ranking.}
\label{tab:decision}
\small
\setlength{\tabcolsep}{4pt}
\begin{tabular}{p{0.23\linewidth}p{0.32\linewidth}p{0.32\linewidth}}
\toprule
Criterion & SAO RL loop & Evo-Harness loop \\
\midrule
Adapted object & Full-parameter $\theta$ & Workspace $\harness$ \\
Observed outcome & 6/6 resolved & 4/6 resolved; 1 verifier failure; 1 timeout \\
Weight-training GPU & 0.567 h (retained window) & None \\
Primary artifact & BF16 checkpoint + value head & One \texttt{prompts/system.md} edit \\
Reversibility & Checkpoint restore & File-level rollback \\
Target bottleneck & Behavior present but weak & Capability present but poorly elicited \\
Evaluation status & Base/combined not evaluated & Base/combined not evaluated \\
\bottomrule
\end{tabular}
\end{table}

\subsection{Why the harness loop is sample efficient}
The harness loop's efficiency claim rests on three implementation properties. First, execution is one-shot at the task level: each task is executed once, and the resulting context is not replayed for every candidate workspace. Second, the reflection signal is targeted: the evidence index exposes rollout records, captures, task files, and Docker listings, and the reflection prompt prioritizes failures and negative feedback. Third, compilation is batch-level and produces concrete file edits: prompts, tools, skills, exemplars, and manifest files can be revised without re-running the original tasks. This avoids both the population of candidate prompts and the repeated task evaluations of prompt-level evolutionary search. These properties connect to reflective methods \citep{shinn2023reflexion,madaan2023selfrefine}; the current Evo-Harness path materializes the result as auditable workspace-file deltas rather than as a new skill for every episode.

\subsection{SAO versus GRPO for agentic tasks}
SAO's single-rollout design is a natural fit for agentic environments. Group-wise methods require multiple rollouts per prompt, which is expensive when each rollout is a multi-turn Docker session; they also introduce a synchronization barrier. We do not import GRPO's group mean/std, \texttt{pass@k} margin, or group-relative length anchor: same-task samples are sparse, repeated SWE rollouts are noisy, and trace length is driven by tool interaction rather than answer sampling. Instead, the completed path uses turn-level SAO advantages, a task-level terminal baseline when multiple samples exist, a critic trained on detached features, and length-adaptive actor $\lambda$ with skip-observation GAE over exact action tokens. The cost is a scalar value head, fixed clipping thresholds, and a strict exact-capture contract; we view this as a reasonable engineering trade-off for long-horizon agent training, not as evidence that the estimator dominates GRPO in general.

\subsection{Evo-Harness versus prompt optimization and session summaries}
GEPA \citep{agrawal2025gepa}, OPRO \citep{yang2024opro}, and EvoPrompt \citep{guo2024evoprompt} optimize prompt strings. In a long-horizon agent, a prompt edit has low amplitude relative to solver stochasticity, and comparing candidates requires repeatedly executing the same expensive task; GEPA-style optimization also starts from a specified seed adapter. Evo-Harness changes the object from a prompt to a grounded workspace delta, and changes validation from repeated candidate rollouts to one execution per task plus batch-level compilation.

A second naive alternative is to let the solver summarize its own session and append the result as a skill. Evo-Harness also uses LLM reflection, but it treats the reflection as a candidate lesson rather than a finished skill. The candidate is grounded in verifier and tool evidence and compiled as concrete workspace-file changes after comparison with the current workspace. The legacy curator implements \textsc{Add}, \textsc{Merge}, \textsc{Revise}, and \textsc{Skip} operations, but the completed run reported here uses the newer workspace-edit path. Direct session summaries are usually self-judged, single-session, and append-only; they cannot distinguish recurring procedures from task-specific artifacts or resolve conflicts with existing files. The advantage of Evo-Harness is therefore not better summarization, but grounded and auditable curation of what becomes reusable guidance.

\subsection{Limitations and threats to validity}
\textbf{No combined or interleaved evaluation.} The completed SAO and Evo-Harness runs use the same task-level interaction pattern but were run separately, and we do not evaluate their post-hoc composition. We also do not measure whether alternating $\theta$ and $\harness$ updates on the same stream helps, or how update scheduling changes the result.

\textbf{Verifier feedback is required.} Evo-Harness needs a reliable external signal for failure and success. In open-ended domains, the verifier may be noisy or gameable, and the evolved harness may overfit to the evaluation distribution. Self-judged session summaries are not a substitute for grounded feedback.

\textbf{Base-harness expressiveness.} Harness evolution edits the workspace that materializes the base harness. Prompts, tool descriptions, skills, and planning templates are easy to revise, but adding executable tool code raises safety and review questions; such edits require stronger sandboxing and human review than text-level workspace changes.

\textbf{Reflection and compilation quality.} The method replaces repeated task execution with reasoning over failures and batches. If the reflection model or compiler misidentifies the cause of failure, a misleading skill can be promoted. The \textsc{Skip} and \textsc{Revise} operations mitigate this but do not eliminate it.

\textbf{No candidate re-execution.} Avoiding candidate re-runs is a compute advantage, but a promoted skill is validated only by future tasks. A stronger gate that does not replay the original task remains open.

\textbf{Run-level failure handling.} The Evo-Harness run contains one 3,000-second execution timeout with no completed episode or final-trace snapshot; we retain it as a failure rather than relabeling it as a model error. The gateway capture nevertheless retains 101 model responses from that task. The 4/6 solve rate is therefore a lower bound under this timeout budget, and the completed-rollout token medians include a zero-token observation.

\textbf{Accounting mismatch.} Completed-rollout API totals, final-turn snapshots, run-level token budgets, and GPU-cost windows are not reconcilable from the retained artifacts. We report each accounting with its source and do not infer a single efficiency winner from them.

\textbf{System-level evidence.} The forward-call reductions in Table~\ref{tab:streaming} are counting and numerical-agreement evidence from a fake CPU model, not wall-clock measurements. The 8GB memory result is one device and one chunk setting; precision/throughput curves for BF16, FP8, and MXFP4 and multi-concurrency 200k serving are not reported. Process-group isolation and restart are implementation designs; we do not claim a measured zero-residue invariant. These limitations keep the systems claims at the level of enabling mechanisms rather than standalone performance or precision conclusions.

\textbf{Evaluation scope.} Our results cover one base agent, one model family, one benchmark, and six SWE-bench Verified instances. We do not evaluate base or combined configurations, and we make no claims about OfficeBench, Terminal-Bench, or cross-domain isolation. The relative advantage of weight updates versus harness evolution may shift with model scale, task horizon, and tool quality.

\textbf{Statistical strength and run continuity.} The solve counts come from one final completed artifact state per specialization loop on six instances, with multiple resume/reflection attempts recorded in the event logs. Evo's three reflection attempts reuse the same six rollouts, and SAO's log records resume and reuse events. We therefore emphasize the process, token, and reasoning audits rather than treating the 6/6 versus 4/6 difference as a broad benchmark result or as a single uninterrupted execution.

\textbf{Safety of self-editing.} An evolved harness can change prompts, tool descriptions, skills, and other workspace files. This is a controlled form of self-modification, and it should be reviewed before deployment, especially when tools have side effects or when the edit adds executable code.

\subsection{Future work}
Several directions follow naturally. First, a held-out retention gate is needed to test the constraint in Eq.~\ref{eq:retention} rather than treating it as a design motivation. Second, an update scheduler could interleave $\theta$ and $\harness$ revisions on the same rollout stream and test whether joint online adaptation works better than separately specialized artifacts. Third, meta-evolution of harness \emph{structure}---adding, removing, or rewiring tools---would extend workspace evolution beyond content edits, with explicit safety review for executable changes. Fourth, cross-adapter composition (e.g., SWE + finance for a fintech agent) requires conflict resolution beyond right-biased override, and a router policy could select specialized workspaces online from the task description. Fifth, the serving stack needs controlled BF16/FP8/MXFP4 comparisons of logprob drift, acceptance rate, throughput, and task quality, plus GPU profiling that separates fusion, prefix reuse, and chunk-size effects. Finally, combining harness evolution with test-time compute scaling \citep{cobbe2021training,lightman2023verify} may allow the workspace to choose reasoning depth per task rather than fixing it globally.

\section{Conclusion}

We studied domain specialization as joint adaptation of policy weights and harness workspace, with systems contributions on both sides. The SAO RL path combines a real-time full-parameter update loop with turn-level action-token credit, exact 200k streaming, fused objectives, prefix reuse, bounded chunking, and restart-based quantized serving. The Evo-Harness path turns one-shot grounded evidence into an auditable workspace edit without policy weight training. On six SWE-bench Verified instances, SAO resolves all six rollouts while Evo-Harness resolves four and produces an interpretable, reversible final edit after three reflection attempts on the same evidence. The process, token, reasoning, and implementation audits show different temporal profiles and cost structures rather than a single efficiency ranking. We do not evaluate the base or combined configurations, other benchmarks, or a held-out retention gate; these are explicit next steps. The broader lesson is that vertical agent specialization should not be framed as a competition between model weights and scaffolding: making both loops observable, composable, and eventually schedulable is the central engineering and research problem.

\section*{Reproducibility statement}
The repository contains the Codev recipes, the workspace/adapter schema, and the evaluation scripts. The completed run directories retain the exact checkpoint paths, resolved configurations, captured traces, and process-reward events. The SAO and Evo-Harness settings are specified in Sections~\ref{sec:setup} and~\ref{sec:system}. Both completed paths record an RTX 4060 Laptop GPU and no policy weight training for Evo; full reproduction still depends on the recorded model server, Docker environment, SWE-bench data, model weights, and dependency versions, and the retained cost ledgers do not cover every resumed session. The streaming forward-call numbers in Table~\ref{tab:streaming} are implementation records rather than independently packaged benchmark artifacts.

\bibliographystyle{plainnat}
\bibliography{ref}

@article{schulman2017ppo,
  title = {Proximal Policy Optimization Algorithms},
  author = {Schulman, John and Wolski, Filip and Dhariwal, Prafulla and Radford,
            Alec and Klimov, Oleg},
  journal = {arXiv preprint arXiv:1707.06347},
  year = {2017},
  url = {https://arxiv.org/abs/1707.06347},
}

@article{shao2024deepseekmath,
  title = {DeepSeekMath: Pushing the Limits of Mathematical Reasoning in Open
           Language Models},
  author = {Shao, Zhihong and Wang, Peiyi and Zhu, Qihao and Xu, Runxin and Song
            , Junxiao and Bi, Xiao and Zhang, Haowei and Zhang, Mingchuan and Li,
            Y. K. and Wu, Y. and Guo, Daya},
  journal = {arXiv preprint arXiv:2402.03300},
  year = {2024},
  url = {https://arxiv.org/abs/2402.03300},
}

@article{hou2026sao,
  title = {Single-Rollout Asynchronous Optimization for Agentic Reinforcement
           Learning},
  author = {Hou, Zhenyu and Li, Yujiang and Tang, Jie and Dong, Yuxiao},
  journal = {arXiv preprint arXiv:2607.07508},
  year = {2026},
  url = {https://arxiv.org/abs/2607.07508},
}

@article{agrawal2025gepa,
  title = {GEPA: Reflective Prompt Evolution Can Outperform Reinforcement
           Learning},
  author = {Agrawal, Lakshya A. and Tan, Shangyin and Soylu, Dilara and Ziems,
            Noah and Khare, Rishi and Opsahl-Ong, Krista and Singhvi, Arnav and
            Shandilya, Herumb and Ryan, Michael J. and Jiang, Meng and Potts,
            Christopher and Sen, Koushik and Dimakis, Alexandros G. and Stoica,
            Ion and Klein, Dan and Zaharia, Matei and Khattab, Omar},
  journal = {arXiv preprint arXiv:2507.19457},
  year = {2025},
  url = {https://arxiv.org/abs/2507.19457},
}

@inproceedings{guo2024evoprompt,
  title = {Connecting Large Language Models with Evolutionary Algorithms Yields
           Powerful Prompt Optimizers},
  author = {Guo, Qingyan and Wang, Rui and Guo, Junliang and Li, Bei and Song,
            Kaitao and Tan, Xu and Liu, Guoqing and Bian, Jiang and Yang, Yujiu},
  booktitle = {International Conference on Learning Representations},
  year = {2024},
  url = {https://arxiv.org/abs/2309.08532},
}

@inproceedings{yang2024opro,
  title = {Large Language Models as Optimizers},
  author = {Yang, Chengrun and Wang, Xuezhi and Lu, Yifeng and Liu, Hanxiao and
            Le, Quoc V. and Zhou, Denny and Chen, Xinyun},
  booktitle = {International Conference on Learning Representations},
  year = {2024},
  url = {https://arxiv.org/abs/2309.03409},
}

@misc{wei2026evoharnesscontexttoharnessskillcompilation,
  title = {Evo-Harness: Context-to-Harness Skill Compilation for Self-Evolving
           Agents},
  author = {Tianxin Wei and Zhan Shi and Minhua Lin and Bing He and Zewen Liu
            and Yisi Sang and Yuanchen Bei and Xuying Ning and Jiaru Zou and
            Ting-Wei Li and Xiao Lin and Yanjun Zhao and Chi Wang and Benoit
            Dumoulin and Dakuo Wang and Jingrui He and Hanqing Lu},
  year = {2026},
  eprint = {2608.15071},
  archivePrefix = {arXiv},
  primaryClass = {cs.AI},
  url = {https://arxiv.org/abs/2608.15071},
}

@inproceedings{yang2024sweagent,
  title = {SWE-agent: Agent-Computer Interfaces Enable Automated Software
           Engineering},
  author = {Yang, John and Jimenez, Carlos E. and Wettig, Alexander and Lieret,
            Kilian and Yao, Shunyu and Narasimhan, Karthik and Press, Ofir},
  booktitle = {Advances in Neural Information Processing Systems},
  year = {2024},
  url = {https://arxiv.org/abs/2405.15793},
}

@inproceedings{yao2023react,
  title = {ReAct: Synergizing Reasoning and Acting in Language Models},
  author = {Yao, Shunyu and Zhao, Jeffrey and Yu, Dian and Du, Nan and Shafran,
            Izhak and Narasimhan, Karthik and Cao, Yuan},
  booktitle = {International Conference on Learning Representations},
  year = {2023},
  url = {https://arxiv.org/abs/2210.03629},
}

@inproceedings{pfeiffer2021adapterfusion,
  title = {AdapterFusion: Non-Destructive Task Composition for Transfer Learning
           },
  author = {Pfeiffer, Jonas and Kamath, Aishwarya and Ruckl{\'e}, Andreas and
            Cho, Kyunghyun and Gurevych, Iryna},
  booktitle = {Proceedings of the 16th Conference of the European Chapter of the
               Association for Computational Linguistics},
  pages = {487--503},
  year = {2021},
  publisher = {Association for Computational Linguistics},
  doi = {10.18653/v1/2021.eacl-main.39},
  url = {https://aclanthology.org/2021.eacl-main.39/},
}

@article{hu2021lora,
  title = {LoRA: Low-Rank Adaptation of Large Language Models},
  author = {Hu, Edward J. and Shen, Yelong and Wallis, Phillip and Allen-Zhu,
            Zeyuan and Li, Yuanzhi and Wang, Shean and Wang, Lu and Chen, Weizhu},
  journal = {arXiv preprint arXiv:2106.09685},
  year = {2021},
  url = {https://arxiv.org/abs/2106.09685},
}

@misc{jimenez2024swebenchlanguagemodelsresolve,
  title = {SWE-bench: Can Language Models Resolve Real-World GitHub Issues?},
  author = {Carlos E. Jimenez and John Yang and Alexander Wettig and Shunyu Yao
            and Kexin Pei and Ofir Press and Karthik Narasimhan},
  year = {2024},
  eprint = {2310.06770},
  archivePrefix = {arXiv},
  primaryClass = {cs.CL},
  url = {https://arxiv.org/abs/2310.06770},
}

@article{wang2024officebench,
  title = {OfficeBench: Benchmarking Language Agents across Multiple
           Applications for Office Automation},
  author = {Wang, Zilong and Cui, Yuedong and Zhong, Li and Zhang, Zimin and Yin
            , Da and Lin, Bill Yuchen and Shang, Jingbo},
  journal = {arXiv preprint arXiv:2407.19056},
  year = {2024},
  url = {https://arxiv.org/abs/2407.19056},
}

@article{merrill2026terminalbench,
  title = {Terminal-Bench: Benchmarking Agents on Hard, Realistic Tasks in
           Command Line Interfaces},
  author = {Merrill, Mike A. and Shaw, Alexander G. and Carlini, Nicholas and
            others},
  journal = {arXiv preprint arXiv:2601.11868},
  year = {2026},
  url = {https://arxiv.org/abs/2601.11868},
}

@inproceedings{liu2024agentbench,
  title = {AgentBench: Evaluating LLMs as Agents},
  author = {Liu, Xiao and Yu, Hao and Zhang, Hanchen and Xu, Yifan and Lei,
            Xuanyu and Lai, Hanyu and Gu, Yu and Ding, Hangliang and Men, Kaiwen
            and Yang, Kejuan and others},
  booktitle = {International Conference on Learning Representations},
  year = {2024},
  url = {https://arxiv.org/abs/2308.03688},
}

@article{chen2025retaining,
  title = {Retaining by Doing: The Role of On-Policy Data in Mitigating
           Forgetting},
  author = {Chen, Howard and Razin, Noam and Narasimhan, Karthik and Chen, Danqi
            },
  journal = {Proceedings of the 43rd International Conference on Machine
             Learning},
  year = {2026},
  eprint = {2510.18874},
  archivePrefix = {arXiv},
  primaryClass = {cs.LG},
  url = {https://arxiv.org/abs/2510.18874},
}

@article{kirkpatrick2017catastrophic,
  title = {Overcoming Catastrophic Forgetting in Neural Networks},
  author = {Kirkpatrick, James and Pascanu, Razvan and Rabinowitz, Neil and
            Veness, Joel and Desjardins, Guillaume and Rusu, Andrei A. and Milan,
            Kieran and Quan, John and Ramalho, Tiago and Grabska-Barwinska,
            Agnieszka and Hassabis, Demis and Clopath, Claudia and Kumaran,
            Dharshan and Hadsell, Raia},
  journal = {Proceedings of the National Academy of Sciences},
  volume = {114},
  number = {13},
  pages = {3521--3526},
  year = {2017},
  doi = {10.1073/pnas.1611835114},
}

@misc{chen2026codev,
  author = {Chen, BoHao},
  title = {Codev: Terminal-Native AI Development Agent},
  year = {2026},
  howpublished = {\url{https://github.com/bhaochen/codev}},
}

@misc{langchain2023,
  author = {LangChain Contributors},
  title = {LangChain: Building Applications with LLMs through Composability},
  year = {2023},
  howpublished = {\url{https://github.com/langchain-ai/langchain}},
}

@misc{autogpt2023,
  author = {AutoGPT Contributors},
  title = {AutoGPT: An Autonomous GPT-4 Experiment},
  year = {2023},
  howpublished = {\url{https://github.com/Significant-Gravitas/AutoGPT}},
}

@misc{anthropic2024claude-code,
  author = {Anthropic},
  title = {Claude Code: AI Pair Programming in Your Terminal},
  year = {2024},
  howpublished = {\url{https://docs.anthropic.com/en/docs/claude-code}},
}

@misc{abgoswam2024swegrpo,
  author = {Abhishek Goswami},
  title = {SWE in Production: GRPO Training on SWE-Bench -- Theory and Practice},
  year = {2024},
  howpublished = {\url{https://github.com/abgoswam/swe_in_prod_vizuara_01}},
  note = {Detailed guide combining theory and practice of one-step GRPO training
          on SWE-Bench},
}

@article{cobbe2021training,
  title = {Training Verifiers to Solve Math Word Problems},
  author = {Cobbe, Karl and Kosaraju, Vineet and Bavarian, Mohammad and Chen,
            Mark and Jun, Heewoo and Kaiser, Lukasz and Plappert, Matthias and
            Tworek, Jerry and Hilton, Jacob and Nakano, Reiichiro and others},
  journal = {arXiv preprint arXiv:2110.14168},
  year = {2021},
}

@article{lightman2023verify,
  title = {Let's Verify Step by Step},
  author = {Lightman, Hunter and Kosaraju, Vineet and Burda, Yura and Edwards,
            Harrison and Baker, Bowen and Lee, Taehoon and Leike, Jan and
            Schulman, John and Sutskever, Ilya and Cobbe, Karl},
  journal = {arXiv preprint arXiv:2305.20050},
  year = {2023},
}

@inproceedings{shinn2023reflexion,
  title = {Reflexion: Language Agents with Verbal Reinforcement Learning},
  author = {Shinn, Noah and Cassano, Federico and Gopinath, Ashwin and
            Narasimhan, Karthik and Yao, Shunyu},
  booktitle = {Advances in Neural Information Processing Systems},
  year = {2023},
  url = {https://arxiv.org/abs/2303.11366},
}

@inproceedings{madaan2023selfrefine,
  title = {Self-Refine: Iterative Refinement with Self-Feedback},
  author = {Madaan, Aman and Tandon, Niket and Gupta, Prakhar and Hallinan,
            Skyler and Gao, Luyu and Wiegreffe, Sarah and Alon, Uri and Dziri,
            Nouha and Prabhumoye, Shrimai and Yang, Yiming and Gupta, Shashank
            and Majumder, Bodhisattwa Prasad and Hermann, Katherine and Welleck,
            Sean and Yazdanbakhsh, Amir and Clark, Peter},
  booktitle = {Advances in Neural Information Processing Systems},
  year = {2023},
  url = {https://arxiv.org/abs/2303.17651},
}

@article{wang2023voyager,
  title = {Voyager: An Open-Ended Embodied Agent with Large Language Models},
  author = {Wang, Guanzhi and Xie, Yuqi and Jiang, Yunfan and Mandlekar, Ajay
            and Xiao, Chaowei and Zhu, Yuke and Fan, Linxi and Anandkumar, Anima},
  journal = {arXiv preprint arXiv:2305.16291},
  year = {2023},
  url = {https://arxiv.org/abs/2305.16291},
}

@inproceedings{schick2023toolformer,
  title = {Toolformer: Language Models Can Teach Themselves to Use Tools},
  author = {Schick, Timo and Dwivedi-Yu, Jane and Dess{\`i}, Roberto and
            Raileanu, Roberta and Lomeli, Maria and Hambro, Eric and Zettlemoyer,
            Luke and Cancedda, Nicola and Scialom, Thomas},
  booktitle = {Advances in Neural Information Processing Systems},
  year = {2023},
  url = {https://arxiv.org/abs/2302.04761},
}

\end{document}